\documentclass[twoside]{article}

\usepackage[accepted]{aistats2024}
\usepackage{amsmath} 
\usepackage{algorithm}
\usepackage{graphicx}
\usepackage{algpseudocode}
\usepackage{amssymb}
\usepackage{mathtools}
\usepackage{amsthm}
\def\Var{{\textrm{Var}}\,}
\DeclareMathOperator*{\argmin}{arg\,min}
\usepackage{hyperref}       % hyperlinks
\usepackage{url}            % simple URL typesetting
\usepackage{booktabs}       % professional-quality tables
\usepackage{amsfonts}       % blackboard math symbols
\usepackage{nicefrac}       % compact symbols for 1/2, etc.
\usepackage{multirow}
\usepackage{placeins}

\usepackage{mathrsfs}

\usepackage{shortbold}

\usepackage[round]{natbib}

\begin{document}



\twocolumn[

\aistatstitle{Approximate Bayesian Class-Conditional Models under Continuous Representation Shift}

\aistatsauthor{ Thomas L.~Lee \And Amos Storkey }

\aistatsaddress{School of Informatics\\
                University of Edinburgh\\
                \texttt{T.L.Lee-1@sms.ed.ac.uk}\\ 
                \And
                School of Informatics\\
                University of Edinburgh\\
                \texttt{a.storkey@ed.ac.uk}\\} ]

\begin{abstract} 
For models consisting of a classifier in some representation space, learning online from a non-stationary data stream often necessitates changes in the representation. So, the question arises of what is the best way to adapt the classifier to shifts in representation. Current methods only slowly change the classifier to representation shift, introducing noise into learning as the classifier is misaligned to the representation. We propose DeepCCG, an empirical Bayesian approach to solve this problem. DeepCCG works by updating the posterior of a class conditional Gaussian classifier such that the classifier adapts in one step to representation shift. The use of a class conditional Gaussian classifier also enables DeepCCG to use a log conditional marginal likelihood loss to update the representation. To perform the update to the classifier and representation, DeepCCG maintains a fixed number of examples in memory and so a key part of DeepCCG is selecting what examples to store, choosing the subset that minimises the KL divergence between the true posterior and the posterior induced by the subset. We explore the behaviour of DeepCCG in online continual learning (CL), demonstrating that it performs well against a spectrum of online CL methods and that it reduces the change in performance due to representation shift.
\end{abstract}

\section{INTRODUCTION} 
\begin{figure*}[t]
\begin{center}
\includegraphics[trim={0cm 5.9cm 0cm 4.9cm}, scale=0.5, clip]{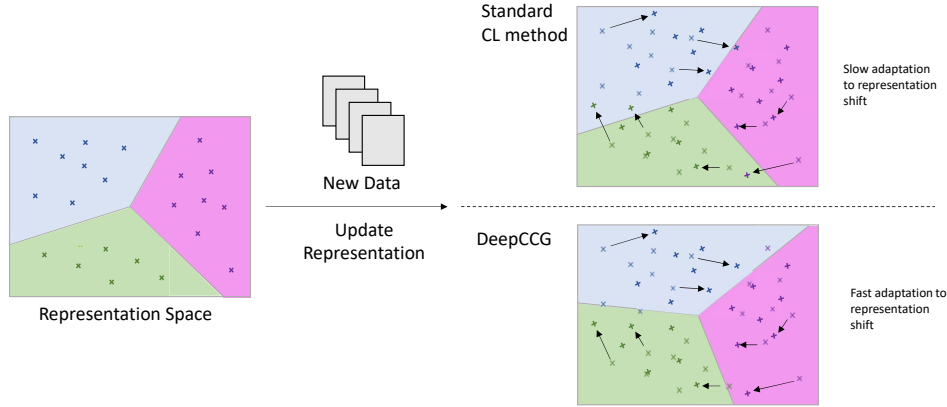}
\end{center}
\vspace{-2mm}
\caption{Diagram showing that when updating a model on new data current online continual learning methods only slowly adapt the classifier, i.e. decision boundary in representation space, to representation shift. On the other hand, DeepCCG quickly adapts the decision boundary, improving learning as the classifier is better matched to the current representation. This is illustrated in the diagram as for DeepCCG the decision boundary is adjusted in a single update such that the shifted representations, which are pointed to by arrows, all remain in the correct regions while for standard CL methods this is not the case.}
\label{fig:RepShift}
%\vspace{-2mm}
\end{figure*}
Currently a large open problem in machine learning is how to update complex neural network models online from a non-stationary data stream \citep{Farquhar2018Towards, antoniou2020defining}. Methods for solving this problem can be seen as a composition of an encoder and classifier---assuming we are looking at classification. The encoder maps data instances to a representation and the classifier given a data instance in representation space assigns a class to it. One of the difficulties encountered is \emph{representation shift} where when updating on new data the representation of old data shifts. To reduce this problem current methods try to minimise representation shift, often by regularising updates using previous data stored in memory \citep{Delange2021A, van2019three}. However, most methods still use a standard classifier (a fully connected layer, then a softmax) when learning, which when trained normally only slowly adapts to representation shift. This means that in the meantime the classifier and representation are missaglined, introducing noise into learning and so potentially making the representation forget more information about previous data than needed (see analysis of representation shift in Section~\ref{sec:repshift}). 

In this paper we propose a clean way to adapt a classifier in one step to representation shift, preventing the misalignment of the classifier and representation. To do this we propose a new method DeepCCG which leverages a Bayesian class-conditional Gaussian model for classification. The use of such a classifier has three main effects. First, the posterior of the parameters of the class-conditional model can be recomputed in one step to adjust to representation shift. This means that the classifier and the shifted representation will be better aligned, removing the potential for the representation to unnecessarily change to better fit the outdated classifier. Second, the use of a Bayesian class-conditional Gaussian classifier, allows for the use of a conditional marginal likelihood for optimizing the parameters of the embedding function. This ensures that the representation learnt is well matched to the classifier. Third, a fixed subset of data is chosen to track representation shift, by selecting the subset that best recreates the posterior distribution. This method for selecting samples is shown to be robust to general kinds of representation shift (see Section~\ref{sample selection}) and is required for DeepCCG to perform well (see ablation study in Section~\ref{Ablation}). 

To examine the behaviour of DeepCCG we look at the online continual learning (CL) setting \citep{Chaudhry2019Efficient, Aljundi2019Gradient}. This is a common setting where a learner sees a non-stationary stream of batches of data. The results of our experiments show that DeepCCG performed best out of all methods tested, highlighting the benefit of quickly adapting to representation shift.

The main contributions of this work are: 
\begin{itemize}
    \item A method DeepCCG, which by learning the classifier in a tractable Bayesian manner adapts to representation shift in one step and performs best in our online CL experiments.

    \item Use of a log conditional marginal likelihood loss term to fit the embedding function, which reduces representation shift and aligns the representation to the class-conditional Gaussian model.

    \item A new method to select what examples to store in memory, by minimising the KL-divergence between the true posterior and the one induced by the subset of data to be stored in memory.
\end{itemize}

\section{RELATED WORK} 
There are three main paradigms for solving online CL problems: \emph{regularisation}, \emph{parameter-isolation} and \emph{replay} \citep{Delange2021A}. Our work is most closely related to \emph{Replay} methods which aim to solve CL problems by storing a subset of previously seen examples, which are then trained on alongside new incoming data. Replay methods have been shown to have competitive if not the best performance across many settings \citep{van2019three, wu2022pretrained, Mirzadeh2020Understanding}. The standard replay method is experience replay (ER) \citep{Chaudhry2020Continual, chaudhry2019tiny, aljundi2019online}, which in the online setting looked at in this work, selects examples to store using reservoir sampling \citep{vitter1985random}, we call this variant ER-reservoir \citep{Chaudhry2020Continual}. One of the main questions to be answered by a replay-based approach is how to select what examples to store in memory. While reservoir sampling has been shown to be very effective \citep{Wiewel2021Entropy} there have been other methods proposed for sample selection, for example ones which use information-theoretic criteria \citep{Wiewel2021Entropy} and others maximising the diversity of the gradients of stored examples \citep{Aljundi2019Gradient}. There has also been a Bayesian method proposed to select samples called InfoGS \citep{Sun2022Information}, which is somewhat similar to our method but only uses the probability model to select samples, not for prediction or training. Additionally, their exists methods to select what examples to replay at each update step \citep{aljundi2019online, shim2020online} which are complementary/orthogonal to this work.  

\:

There has been considerable work on using Bayesian methods in CL \citep{kessler2023sequential, Ebrahimi2020Uncertainty-guided, Kurle2020Continual, lyu2023overcoming}, perhaps inspired by the fact that true online Bayesian inference cannot suffer from catastrophic forgetting \citep{nguyen2017variational}. Bayesian perspectives have mainly been used for regularisation based methods, where a popular approach is to use variational inference \citep{nguyen2017variational, farquhar2019unifying, zeno2018task}. These variational inference methods focus on the offline CL setting where a learner has access to all of the data of a task at the same time and, in previous work, are often limited to being used in conjunction with small neural networks, mainly due to the need to sample multiple networks when calculating the loss \citep{Henning2021Posterior, nguyen2017variational}. Therefore, these methods are not suited to the settings we consider in this paper. Bayesian methods have also been used in generative replay based approaches, where instead of storing and replaying real samples they use generated pseudo-samples \citep{Rao2019Continual}. Generative replay methods focus on the offline scenario, where it is possible to iterate over the whole of a task's data to fit a generator, while we look at the more realistic online scenario in this work. Finally, when it comes to replay with real examples, there has been relatively little work on using Bayesian methods, which we aim to help to fill in by proposing DeepCCG.

A closely related work to DeepCCG is iCaRL \citep{rebuffi2017icarl}. It is similar because both methods use a class-conditional Gaussian classifier. However, iCaRL only uses a class-conditional Gaussian classifier at test time, using a per-class sigmoid classifier in training. Therefore, iCaRL does not solve the problem we set out to address, that of quickly adapting to representation shift in training, as its classifier in training is similar to other standard CL algorithms and only slowly adapts to representation shift. We also note that while iCaRL's sample selection mechanism is different from DeepCCGs, it can be seen to approximate the same general objective---selecting the subset which best recreates the mean of the whole data, in representation space. Therefore, our principled Bayesian derivation of DeepCCG's sample selection mechanism can be applied to iCaRL's giving it a theoretical grounding, which is not looked at in the iCaRL paper. Another similar approach to DeepCCG is ProtoCL \citep{kessler2023sequential}, which appeared after DeepCCG's initial release. It is similar in that it uses a class-conditional Gaussian classifier but the learning of the classifier and representation is different and ProtoCL uses random sampling to select what examples to store in memory, which for DeepCCG is shown to have bad performance (see Table~\ref{tab:Ablation}). 

\section{ONLINE CL}
\begin{figure*}[t] %shoud these figs be placed futher down in the text?
\begin{center}
\includegraphics[trim={4.8cm 7.7cm 6cm 5cm}, scale=0.6, clip]{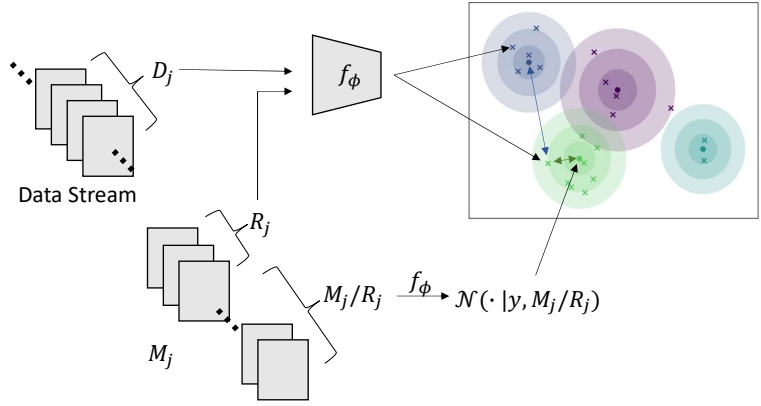}
\end{center}
\vspace{-2mm}
\caption{Diagram of DeepCCG's training routine. At time $j$ the learner is given a sample of data $B_j$ and has a memory of stored datapoints $M_j$. The memory is randomly split into replay data $R_j$ and the rest $M_j/R_j$. Learning happens by taking a gradient step on the parameters of the embedding function $\phiB$ using a log conditional marginal likelihood function over $B_j$ and $R_j$, where $M_j/R_j$ is used to induce a posterior over the means of the per-class Gaussians and so define the conditional marginal likelihood function used. Therefore, training aims to move the data points into per-class clusters, by drawing the embedded examples of $B_j$ and $R_j$ towards their own class means and away from the other class means, as shown by the coloured arrows in the figure.}
\label{fig:DeepCCG}
\end{figure*}
This work uses the Online CL setting to study how to deal with representation shift; where a learner sees a non-stationary data stream consisting of batches of data. Specifically, we consider here classification problems. Let $X$ denote a shared data space, and $C$ denotes the set of all classes being considered. Additionally, let $t \in T$ denote a particular task. A task $t$ is a data generator which generates data such that each each example consists of a data instance $\xB\in X$ and associated label $y\in C_t$. Also, $C_t \subseteq C$ are the subset of classes which all the data generated by task $t$ must belong to.  

During training, the learner receives a temporal sequence of data batches with their task identifiers, $((B_j,t_{B_j})|j=1,2,\ldots,N)$. Each batch $B_j$ consists of a set of examples drawn from the same task. The task identifier $t_{B_j}$ denotes what task that is, effectively indicating the classes which a data instance $\xB$ of that task can belong to. The learner trains on each batch in turn and after training on a batch it must discard all the data it does not store in a fixed-size memory buffer $M$. Also, the learner never revisits the same batch again. 

At test time, we consider two evaluation scenarios, task-incremental and class-incremental learning. For both scenarios the learnt model is evaluated on an unseen test set from all the classes seen during training. For task-incremental learning the learner is provided with task identifiers for each test data instance and without in class-incremental learning. This experimental setup is commonly used in the continual learning literature \citep{Chaudhry2019Efficient}. Additionally, we provide a reference table of terms used in online CL and defined in this section in Appendix~\ref{appen:CLterms}.

\section{DEEP CLASS-CONDITIONAL GAUSSIANS}
We propose an empirical Bayesian method DeepCCG, which can adapt quickly to representation shift. DeepCCG consists of three main parts. Firstly, a Bayesian class-conditional Gaussian classifier is used on top of a neural network embedding function. The classifier is fit by recomputing its posterior whenever there is a shift in representation; which only requires a single forward pass through the neural network embedding function. Secondly, by using a class-conditional Gaussian classifier we can learn the embedding function using a log conditional marginal likelihood loss. The loss is derived from and uses the learnt classifier in an end-to-end manner and aims to fit a representation that best allows for the classifier to discriminate between classes. Additionally, by conditioning the marginal likelihood on some stored data, the loss aims to minimise the representation shift of previously seen classes. Finally, DeepCCG uses a new method based on minimising information loss to select a subset of examples to store in memory, these examples are needed when updating both the classifier and the embedding function, to model and reduce representation shift. The sample selection mechanism is shown to be robust to certain kinds of representation shift. These three parts are described in detail below and additionally a pseudocode description of the learning of DeepCCG is presented in Algorithm~\ref{alg:deepccg}.
\begin{algorithm}[t]
    \centering
    \caption{DeepCCG update step at time $j$}\label{algorithm1}
    \footnotesize
    \begin{algorithmic}
    \State \textbf{input} $B_j$ (training batch), $t_{B_j}$ (task identifier for batch), $M_j$ (current memory), $f_{\phiB_j}$ (embedding function)
    \State
    \State \textbf{Update class-conditional Gaussian classifier and embedding function:}
    \State $R_j = \mathrm{UniformSample}(M_j)$
    \State Calculate posterior of class-conditional Gaussian classifier: $p(\muB_c| M_j/R_j)$ for each class $c$, using Eq.~\ref{eq:post}
    \State Calculate $\log(p(y| \zB = f_{\phiB_j}(\xB), t_{(\xB,y)}, M_j/R_j))$, for each $(\xB,y) \in B_j \cup R_j$, using Eq.~\ref{eq:PPD} and posteriors $p(\muB_c| M_j/R_j)$
    \State Update embedding function: $\phiB_{j+1} = \phiB_{j} + \eta \nabla_{\phiB} \sum_{(x,y) \in B_j \cup R_j} \log(p(y| \zB = f_{\phiB_j}(\xB), t_{(\xB,y)}, M_j/R_j))$
    \State
    \State \textbf{Update memory buffer:}
   \For{each class $c$ in $M_j$} 
    \State initialise $\betaB$
    \For{$1$ to $B$}
    \State $\betaB \gets \betaB + \eta \nabla_{\betaB} \mathcal{L}(\betaB; B_{j, c}, M_{j,c})$, where
    \State  \: $\mathcal{L}(\betaB; B_{j, c}, M_{j,c})$ is defined in Eq.~\ref{eq:SS}
    \EndFor
    \State Set $M_{j+1}$ to be the set of examples, including their task identifiers, with the $m$ largest values in $\betaB$
    \EndFor
    \end{algorithmic}
    \label{alg:deepccg}
\end{algorithm}

\textbf{Probabilistic Classifier} \: We use a Bayesian class-conditional Gaussian model as the classifier for DeepCCG. This model is defined as follows. Let $Z$ denote a representation space and assume that we have a neural network embedding function $f_{\phiB}:X\rightarrow Z$. We define a class-conditional Gaussian model in the representation space $Z$,
\begin{align}
 y | t &\sim  \mathrm{Cat}(C_t, (1/|C_t|) \mathbf{1})\\ 
 \zB | y &\sim \mathcal{N}(\muB_y, \mathbf{I}) \\
 \muB_y &\sim \mathcal{N}(\mathbf{0}, \mathbf{V}_0),
\end{align}
where $t$ is a given task identifier, $\zB=f_{\phiB}(\xB)$, $\{\muB_y|y \in C\}$ are the parameters of the classifier being the per-class means in $Z$ and $\mathrm{Cat}(C_t, (1/|C_t|) \mathbf{1})$ is the uniform categorical distribution over the classes of task $t$. We choose $\mathbf{V}_0=a \mathbf{I}$, where in practice we take $a \rightarrow \infty$. We do not specify a model where the covariance matrix is also a parameter as we learn the embedding function, hence the case of a global shared covariance is implicitly covered through learning a linear remapping to a fixed covariance, meaning we do not lose flexibility by assuming fixed covariances. Also, we assume the labels are conditionally independent given the task, the per-class means $\{\muB_c|c\in C\}$ and the data instances (each an element of $X$) and that $\zB$ is independent of the task given its class $y$. 

The classifier is defined as the posterior predictive distribution of the model, i.e. $p(y|\zB=f_{\phiB}(\xB), t_{\xB}, D)$ given the data instance $\xB$, its task identifier $t_{\xB}$ and some previously seen data $D$. The posterior predictive distribution can be calculated by using the equivalence, 
\begin{multline} 
p(y|\zB=f_{\phiB}(\xB), t_{\xB}, D) = \\ \int p(y|\zB=f_{\phiB}(\xB), t_{\xB}, \thetaB=\{\muB_c|c\in C\})p(\thetaB| D) d\thetaB, \label{eq:intpostpred}
\end{multline}
where
\begin{multline}
p(y|\zB=f_{\phiB}(\xB), t_{\xB}, \thetaB = \{\muB_c|c\in C\}) = \\ \frac{p(\zB|y,\muB_y)p(y|t_{\xB})}{\sum_{c \in C} p(\zB|c,\muB_{c})p(c|t_{\xB})},
\end{multline}
and $p(\thetaB| D)$ is the product over the posteriors $p(\muB_c| D)$ for each class, where
\begin{align} 
    p(\muB_c| D) &= p(\muB_c |D^{Z}_c) \\ 
                    &= \mathcal{N}\left(\muB_y; \overline{D^{Z}_c}, \frac{1}{|D^{z}_c|}\mathbf{I}\right). \label{eq:post}
\end{align}
$D^{Z}_c =\{f_{\phiB}(\xB)|(\xB, y)\in D \land y = c \}$ are the representations of the data instances in $D$ of class $c$ and we use the notation $\overline{S}$ to denote the mean of the elements of a set $S$. The integral in Eq.~\ref{eq:intpostpred} can be computed, giving a closed form expression for the classifier output (see Appendix~\ref{appen:learning} for the full derivation)
\begin{multline}
    p(y|\zB=f_{\phiB}(\xB),t_{\xB}, D) = \\ \frac{\mathcal{N}\left(\zB; \overline{D^{Z}_y}, (1+\frac{1}{|D^{Z}_y|})\mathbf{I} \right)}{\sum_{c \in C_{t_{\xB}}} \mathcal{N}\left(\zB; \overline{D^{Z}_y}, (1+\frac{1}{|D^{Z}_y|})\mathbf{I} \right)}. \label{eq:PPD}
\end{multline}
This shows that the classifier gives the probability of $\xB$ belonging to a class $y$ by how close its representation $\zB=f_{\phiB}(\xB)$ is to the class mean $\overline{D^{Z}_y}$ and by how many examples DeepCCG has already seen of a class, i.e. $|D^{Z}_y|$.   

To fit the classifier, DeepCCG only needs to compute the posterior of the per-class means. This is because this posterior can be used to create the posterior predictive distribution of any data instance. The posterior is tractable and easy to compute, as shown in Eq.~\ref{eq:post}. This is due to our choice of model which is the main reason we use this model along with its good performance when used on top of neural network embeddings \citep{ostapenko2022foundational, hayes2020lifelong}. If $f_{\phiB}$ is fixed, so there is no representation shift, we can update the posterior of the per-class means on seeing batch $B_j$ and its task identifier $t_{B_j}$ by simply using Bayes rule, calculating the posterior $p(\{\muB_c|c\in C\}|B_j, B_{<j})$, where $B_{<j}$ is the union of the batches seen before $B_j$. However, in our setting, we need to update $f_{\phiB}$ continually and so there is representation shift, which changes the value of $p(\{\muB_c|c\in C\}|B_j, B_{<j})$. We cannot recompute $p(\{\muB_c|c\in C\}|B_j, B_{<j})$ as we are unable to store all the previously seen examples. Therefore, DeepCCG stores in memory a buffer of representative examples $M$ and their task identifiers with which we can compute an approximate posterior $p(\{\muB_c|c\in C\}|M)$ after a change in $f_{\phiB}$. It is important to note that we can compute this posterior using a single forward pass through $f_{\phiB}$ and so we can adapt the classifier in one step to representation shift, improving learning as the classifier and representation are never misaligned.  

\textbf{Learning the Embedding} \: DeepCCG uses a novel log conditional marginal likelihood loss term update the embedding function. The log conditional marginal likelihood loss is given by the Bayesian classifier and uses the classifier so that the loss fits a representation which enables the classifier to best separate the classes. At time $j$, in response to the arrival of batch $B_j$, with task identifier $t_{B_j}$, and having the memory buffer $M_j$, DeepCCG takes the following steps to update the current embbeding function $f_{\phiB_{j}}$ (shown in Figure~\ref{fig:DeepCCG}). First, it selects a random set $R_j \subset M_j$, of size $r$, from the memory buffer to replay. Then it calculates the output of the classifier for each example $(\xB,y) \in B_j \cup R_j$ in the current batch and replay set. The classifier output is the probability of the class being $y$ for $\xB$ given by the posterior predictive distribution $p(y|\zB = f_{\phiB_j}(\xB),t_{(\xB,y)}, M_j/R_j)$, where we condition on the rest of the stored examples $M_j/R_j$ and $t_{(\xB,y)}$ is the task identifier for the example. The posterior predictive distributions are calculated using Eq.~\ref{eq:PPD} (where in this case $D=M_j/R_j$). Then the posterior predictive distributions are used as individual log conditional marginal likelihood terms. This leads to the embedding update rule:  
\begin{multline}
    \phiB_{j+1} = \phiB_{j} + \\ \eta \nabla_{\phiB} \sum_{(\xB,y) \in B_j \cup R_j} \log(p(y|\zB = f_{\phiB_j}(\xB), t_{(\xB,y)}, M_j/R_j)).
\end{multline}
DeepCCG conditions the marginal likelihood on some stored data instead of using, as standard, only the marginal likelihood. The reason for this is that conditioning the likelihood on some examples $M_j/R_j$ stabilises the output of the classifier and hence the loss term used. This is because by conditioning the marginal likelihood on $M_j/R_j$, the posterior DeepCCG averages over is more informative than using the unconditioned prior, having some belief of where the position of embedded data instances should be and so provides a better signal to fit the embedding function. Additionally by replaying the examples $R_j$, treating them like new data, the loss leverages a new type of replay, where performing replay is widely known to be effective at minimising unnecessary representation shift \citep{wu2022pretrained}.

\label{sample selection}
\textbf{Sample Selection} \: A key component to DeepCCG is how to select samples to store in the memory buffer. We focus on ensuring the least amount of information is lost about the position of the per-class means of the class-conditional Gaussian classifier, after a shift in representation. Hence, because all the information available to inform the value of the classifiers per-class mean parameters is encapsulated in the full posterior over the seen data, we target storing a set of examples that best recreate that posterior, preventing as much information loss as possible. Therefore, to perform sample selection we minimise the KL divergence between two posterior distributions: the posterior over parameters induced by the new memory being optimized, and the posterior induced by the current batch and the old memory. We keep the number of examples in memory for each class balanced, so minimising the KL divergence is equivalent to minimising each per-class KL divergence. Therefore, we select the new memory for each class using
\begin{multline}
    M_{j+1, y} = \argmin_{M^{\prime}_y}(\mathrm{KL}(p(\muB_y|B_{j, y},M_{j, y}) \lVert p(\muB_y|M^{\prime}_y))) \\ = \argmin_{M^{\prime}_y}(\lVert \overline{B_{j, y}^Z \cup M_{j, y}^Z}-\overline{M_y^{\prime Z}} \rVert^2_2), \label{eq:memmin}
\end{multline}
where $M^{\prime}_y \subseteq B_{j, y} \cup M_{j, y}$, $M_{j+1, y}$ is the new memory to be selected for class $y$, $M_{j, y}$ is the current memory for class $y$ and $B_{j,y}$ is the set of examples of class $y$ in the current batch. Eq.~\ref{eq:memmin} shows that DeepCCG selects the data to store such that they have as close to the same mean as the old memory plus current batch as possible. Importantly, this sample selection mechanism is robust to representation shift, when modeled as an additive i.i.d. change in representation. This is because, given the shifted representations $\zB^{*}= \zB +\epsilonB_{\zB}$, where $\zB \in D^Z \cup M^Z$ and $\epsilonB_{\zB}$ are i.i.d. sampled from an arbitrary distribution with bounded mean and variance, we have that (proved in Appendix~\ref{ProofOfRepShift}):
\begin{multline}
    \mathbb{E}[\mathrm{KL}(p(\muB_y|D^{Z^*}_{j, y},M^{Z^*}_{j, y}) \lVert p(\muB_y|M^{\prime Z^*}_y)))] = \\ \lVert \overline{B_{j, y}^Z \cup M_{j, y}^Z}-\overline{M_y^{\prime Z}} \rVert^2_2 + \nu (\frac{1}{|M^{\prime Z}_{y}|}-\frac{1}{|B_{j, y}^Z \cup M_{j, y}^Z|}),
\end{multline}
where the expectation is taken over the additive shift terms $\epsilonB$, $\nu$ is the sum over the per-dimension variances of the shift distribution, $D^{Z^*}_{j, y} =\{\zB^{*} = \zB + \epsilonB_{\zB} |\zB \in D^{Z}_{j, y}\}$ and $M^{Z*}_{j, y}$ is defined likewise. Therefore, in expectation, the examples selected to be stored in memory after a shift in representation are still the best examples to store in memory, in terms of preserving posterior information. Hence, our sample selection mechanism is robust to this type of representation shift.

Performing the minimization in Eq.~\ref{eq:memmin} is computationally hard, so we utilise a relaxation of the problem using \emph{lasso} \citep{hastie2009elements}, whereby our method selects the new memory by assigning to each embedded input $z_i$ a zero-to-one weight $\beta_i$ and performing gradient decent on the loss 
\begin{multline} 
\mathcal{L}(\betaB; B_{j, y}, M_{j, y}) = \lambda \lVert \betaB \rVert_1 +\\ \Bigg\lVert \overline{B_{j, y}^Z \cup M_{j, y}^Z}- \frac{1}{\lVert \betaB \rVert_1}\sum_{\stackrel{i|(\xB_i, y_i) \in}{B_{j, y} \cup M_{j, y}}} \beta_i z_i \Bigg\rVert^2_2. \label{eq:SS}
\end{multline}   
Then, after termination, our method selects the $m$ examples with the largest weights in $\betaB$ to be the examples stored in memory, where they are stored along with their task identifiers.

\section{EXPERIMENTS}
\label{setup}
\textbf{Benchmarks} \: In our experiments we look at task and class incremental learning in both the disjoint tasks and shifting window settings. Furthermore, we consider three different datasets: CIFAR-100 \citep{krizhevsky2009learning}, MiniImageNet \citep{vinyals2016matching} and CIFAR-10 \citep{krizhevsky2009learning}, where both CIFAR-100 and MiniImageNet contain 100 classes, while CIFAR-10 contains 10 classes. For disjoint tasks, we split data evenly across a certain number of tasks while assigning all examples with a particular class to only one task---this is often called the `split tasks' setting in previous work \citep{Delange2021A, Chaudhry2019Efficient}. We split CIFAR-10 into 5 tasks where there are 2 classes per task and for CIFAR-100 and MiniImageNet we split the dataset into 20 tasks with 5 classes per task. In the alternative shifting window setting, we split the datasets up into tasks by fixing an ordering of the classes $c_1,\ldots, c_k$ and construct the $i$th task by selecting a set of examples from classes $c_i,\ldots,c_{i+l}$, where $l$ is the window length. No two tasks contain the same example and each task has the same number of examples per-class. For CIFAR-10 we use a window length of $2$ and for CIFAR-100 and MiniImageNet we use a window length of $5$. The shifting window setting is somewhat like the blurry task setting considered in other work \citep{Bang2021Rainbow}, where they have in common that there is class overlap between tasks. However, in the shifting window setting there is also temporal locality between the classes seen. Additionally, for all experiments we train with $500$ examples per-class.

We evaluate the methods using a standard metric for CL, average accuracy \citep{Chaudhry2019Efficient}. The average accuracy of a method is the mean accuracy on a reserved set of test data across all tasks after training on all tasks.

\begingroup
\setlength{\tabcolsep}{2.5pt} % Default value: 6pt
\begin{table*}[t]
  \caption{Results of task-incremental and class-incremental learning experiments on CIFAR-10, CIFAR-100 and MiniImageNet, where SW and DT stand for the shifting window and disjoint tasks settings, respectively. We report mean average accuracy with their standard errors across three independent runs. The results show that DeepCCG performs the best out of the methods tested.}
  \label{tab:mainResults}
  \centering
  \begin{tabular}{llllllll}
    \toprule
    & & \multicolumn{2}{c}{CIFAR-10} & \multicolumn{2}{c}{CIFAR-100} & \multicolumn{2}{c}{MiniImageNet} \\
    \cmidrule(r){3-4} \cmidrule(r){5-6} \cmidrule(r){7-8}
    Scenario & Method & SW & DT & SW & DT & SW & DT   \\
    \midrule
    \multirow{ 13}{*}{Task-Inc.} & EWC & $58.68_{\pm1.88}$ & $63.33_{\pm0.87}$ & $36.65_{\pm1.07}$  & $42.39_{\pm0.79}$ & $32.42_{\pm1.13}$ & $29.57_{\pm0.69}$  \\
    & PackNet & $69.29_{\pm2.27}$ & $66.97_{\pm1.47}$ & $40.21_{\pm0.96}$ & $50.28_{\pm0.58}$ & $34.15_{\pm1.28}$ & $37.86_{\pm1.71}$ \\
    & ER-reservoir & $70.44_{\pm0.81}$ & $66.71_{\pm0.90}$ & $54.05_{\pm0.63}$ & $58.31_{\pm1.08}$ & $41.04_{\pm1.54}$ & $40.29_{\pm1.08}$ \\
    & A-GEM & $57.63_{\pm2.16}$ & $57.28_{\pm2.61}$ & $29.01_{\pm1.45}$ & $39.00_{\pm0.75}$ & $26.97_{\pm1.26}$ & $30.08_{\pm1.86}$ \\
    & EntropySS & $67.93_{\pm0.63}$ & $64.96_{\pm0.91}$ & $51.80_{\pm0.70}$  & $56.75_{\pm0.81}$ & $40.03_{\pm0.61}$  & $41.12_{\pm0.51}$ \\
    & GSS & $71.55_{\pm1.45}$ & $67.30_{\pm1.27}$ & $48.20_{\pm0.33}$  & $49.92_{\pm0.50}$ & $37.91_{\pm0.49}$  & $38.77_{\pm0.98}$ \\
    & ER-ACE & $71.01_{\pm0.81}$ & $68.94_{\pm0.24}$ & $52.65_{\pm0.09}$ & $54.57_{\pm0.61}$ & $40.06_{\pm0.56}$ & $39.42_{\pm0.20}$ \\
    & DER++ & $70.86_{\pm1.24}$ & $67.79_{\pm0.84}$ & $53.92_{\pm1.05}$ & $57.08_{\pm0.79}$ & $41.22_{\pm0.35}$ & $41.95_{\pm1.19}$ \\
    & ESMER & $71.00_{\pm0.67}$ & $65.13_{\pm1.92}$ & $53.58_{\pm0.21}$ & $56.90_{\pm0.21}$ & $41.09_{\pm0.93}$ & $42.46_{\pm0.58}$ \\
    & iCaRL & $63.53_{\pm0.12}$ & $67.98_{\pm0.90}$ & $31.77_{\pm0.33}$ & $35.99_{\pm0.55}$ & $30.96_{\pm0.75}$ & $30.47_{\pm1.29}$  \\ 
    & DeepCCG (ours) & $\mathbf{74.65_{\pm2.00}}$ & $\mathbf{69.29_{\pm0.89}}$ & $\mathbf{56.62_{\pm0.29}}$  & $\mathbf{60.46_{\pm0.24}}$ & $\mathbf{42.18_{\pm0.45}}$ & $\mathbf{43.04_{\pm0.64}}$ \\
    \cmidrule(r){2-8}
    & SGD & $63.21_{\pm2.09}$ & $63.58_{\pm0.31}$ & $35.63_{\pm1.27}$  & $42.50_{\pm1.40}$ & $33.23_{\pm0.67}$ & $31.61_{\pm0.70}$ \\
    & Multi-Task (UB) & $96.20_{\pm0.69}$ & $73.37_{\pm1.82}$ & $90.42_{\pm1.87}$ & $59.76_{\pm0.49}$ & $89.81_{\pm0.25}$ & $48.46_{\pm1.21}$ \\
    \midrule
    \midrule
    \multirow{ 12}{*}{Class-Inc.} & EWC & $14.36_{\pm1.64}$ & $13.85_{\pm2.17}$ & $3.77_{\pm0.39}$  & $4.51_{\pm0.19}$ & $2.73_{\pm0.21}$ & $2.84_{\pm0.19}$  \\
    & ER-reservoir & $18.96_{\pm1.18}$ & $16.29_{\pm0.75}$ & $7.76_{\pm0.91}$ & $7.19_{\pm0.56}$ & $5.93_{\pm1.05}$ & $5.15_{\pm0.22}$ \\
    & A-GEM & $11.98_{\pm0.82}$ & $14.88_{\pm0.52}$ & $1.92_{\pm0.04}$ & $3.01_{\pm0.13}$ & $1.21_{\pm0.16}$ & $1.83_{\pm0.04}$ \\
    & EntropySS & $8.93_{\pm1.11}$ & $15.69_{\pm0.78}$ & $5.85_{\pm0.47}$  & $7.87_{\pm0.71}$ & $2.93_{\pm0.45}$  & $5.63_{\pm0.40}$ \\
    & GSS & $16.11_{\pm1.23}$ & $17.61_{\pm0.30}$ & $6.34_{\pm0.20}$  & $5.18_{\pm0.44}$ & $4.52_{\pm0.25}$  & $4.84_{\pm0.16}$ \\
    & ER-ACE & $23.76_{\pm1.61}$ & $20.77_{\pm1.55}$ & $13.51_{\pm0.34}$ & $14.72_{\pm0.32}$ & $8.78_{\pm0.48}$ & $8.29_{\pm0.33}$ \\
    & DER++ & $18.81_{\pm1.17}$ & $15.37_{\pm1.42}$ & $9.53_{\pm0.35}$ & $7.74_{\pm0.17}$ & $6.19_{\pm0.19}$ & $5.78_{\pm0.10}$ \\
    & ESMER & $17.07_{\pm0.89}$ & $17.88_{\pm0.80}$ & $9.04_{\pm0.55}$ & $7.93_{\pm0.14}$ & $6.45_{\pm0.15}$ & $5.77_{\pm0.12}$ \\
    & iCaRL  & $19.48_{\pm0.63}$ & $19.96_{\pm0.80}$ & $4.39_{\pm0.12}$ & $5.02_{\pm0.19}$ & $3.31_{\pm0.11}$ & $3.34_{\pm0.04}$ \\
    & DeepCCG (ours) & $\mathbf{26.87_{\pm0.47}}$ & $\mathbf{24.27_{\pm0.44}}$ & $\mathbf{18.00_{\pm0.72}}$  & $\mathbf{17.32_{\pm0.27}}$ & $\mathbf{11.71_{\pm0.39}}$ & $\mathbf{11.88_{\pm0.71}}$ \\
    \cmidrule(r){2-8}
    & SGD & $13.59_{\pm1.76}$ & $13.76_{\pm1.22}$ & $4.09_{\pm0.14}$  & $3.93_{\pm0.29}$ & $1.28_{\pm0.06}$ & $2.14_{\pm0.22}$ \\
    & Multi-Task (UB) & $40.57_{\pm2.33}$ & $40.57_{\pm2.33}$ & $20.22_{\pm0.42}$ & $20.22_{\pm0.42}$ & $12.41_{\pm0.69}$ & $12.41_{\pm0.69}$ \\
    \bottomrule
  \end{tabular}
  %\vspace{-3pt}
\end{table*}
\endgroup
\textbf{Methods} \: We compare DeepCCG to representative methods of the main paradigms of CL. For regularisation methods we compare against a fixed memory variant of EWC \citep{huszar2018note, kirkpatrick2017overcoming}; for parameter-isolation methods we compare to PackNet \citep{mallya2018packnet}, but only in the task-incremental setting as it requires task identifiers at test time. For replay methods, which includes DeepCCG, we compare against ER-Reservoir \citep{Chaudhry2020Continual}, A-GEM \citep{Chaudhry2019Efficient}, EntropySS \citep{Wiewel2021Entropy}, GSS \citep{Aljundi2019Gradient}, ER-ACE \citep{caccia2021new}, DER++ \citep{buzzega2020dark}, ESMER \citep{sarfraz2022error} and iCaRL \citep{rebuffi2017icarl}. A description of each method compared against is given in Appendix~\ref{appen:methods} and we note here that EntropySS and GSS are memory sample selection strategies for ER. When training, all methods are given task identifiers which state what classes are in a task. Task identifiers are also given at test time in the task-incremental scenario while in class-incremental learning the task-identifier at test time is treated as stating a data instance can belong to any class. We also compare against two baselines: SGD which is when learning is performed using SGD with no modification and Multi-Task which is an upper bound and is when we learn the base neural network offline---with task identifiers for the task-incremental experiments and without for the class-incremental experiments. All methods use the same embedding network for all experiments which is a ResNet18 with six times fewer filters and Instance Normalisation \citep{ulyanov2016instance} instead of Batch Normalisation layers \citep{Ioffe2015Batch}, which is similar to other work \citep{Mirzadeh2020Understanding, farajtabar2019orthogonal}. A batch size of $10$ is used throughout, with all replay methods having a replay size of $10$. A memory size of $10$ examples per-class is used for the task-incremental setting, and $30$ for the harder class-incremental setting. Further experimental details are provided in Appendix~\ref{appen:addExpDetails}.

\subsection{Results}
In the task-incremental experiments, we see from Table~\ref{tab:mainResults} that DeepCCG performs the best out of the methods compared. For example, DeepCCG on CIFAR-100 achieves mean average accuracies of $56.62\%$ and $60.46\%$ for the shifting window and disjoint tasks settings, respectively, which are $2.57\%$ and $2.15\%$ better than any other method. The next best performing methods are ER-reservoir, ER-ACE, ESMER and DER++, which all achieve a second best performance in one of the dataset and setting combination. However there is not one single method that consistently performs second best across all datasets and settings. Overall, DeepCCG had an average performance improvement of $1.62\%$ over the other methods in the task-incremental experiments. Therefore, our experiments show that DeepCCG performs well in task-incremental learning where it is hard to get large performance improvements \citep{prabhu2020gdumb}.    

For the experiments on class-incremental learning, we see that DeepCCG performs well, outperforming the other methods. The results of the class-incremental experiments are shown in Table~\ref{tab:mainResults} and displays for instance that DeepCCG on CIFAR-100 achieves $4.49\%$ and $2.6\%$ better than any other method, for the shifting window and disjoint tasks settings, respectively. Overall, DeepCCG across all datasets and settings in class-incremental learning has an average performance improvement of $3.37\%$, which shows that the method performs well and with the results on task-incremental learning shows that DeepCCG is effective in both scenarios. The next best method was ER-ACE which was consistently the second best method in the class-incremental experiments. Also, it is interesting to note that DeepCCG performs better, in terms of performance improvement over the other methods, in the class-incremental experiments than the task-incremental experiments, achieving an average performance improvement of $1.62\%$ in task-incremental learning and $3.37\%$ in class incremental learning. This finding is surprising as in DeepCCG there is no interaction between the mean parameters of classes which do not appear together in a training task. Therefore, these non-interacting mean parameters could be close together, harming performance in class-incremental learning, but this does not seem to be the case in practice and improving upon this is a direction for future work. 

The results on the shifting window setting show that current methods do not exploit the increased task overlap in the setting and hence do not utilize the increased ability to transfer knowledge between tasks. This is shown in the results as the performance of the methods are very similar between the shifting window and disjoint tasks settings; while, in the task-incremental experiments, the difference between the multi-task upper bound's performance and the methods is much larger in the shifting window setting than that for the disjoint tasks setting.\footnote{For class-incremental experiments the upper bound does not use task identifiers so is identical for the shifting window and disjoint tasks settings.} This suggests that in settings of increased task overlap, where between-task knowledge transfer is more beneficial, current methods do not achieve accuracies near to that which is possible.

\begin{figure}[t]
    \centering
    \includegraphics[trim={0.8cm 0.0cm 2cm 1.5cm}, clip, scale=0.4]{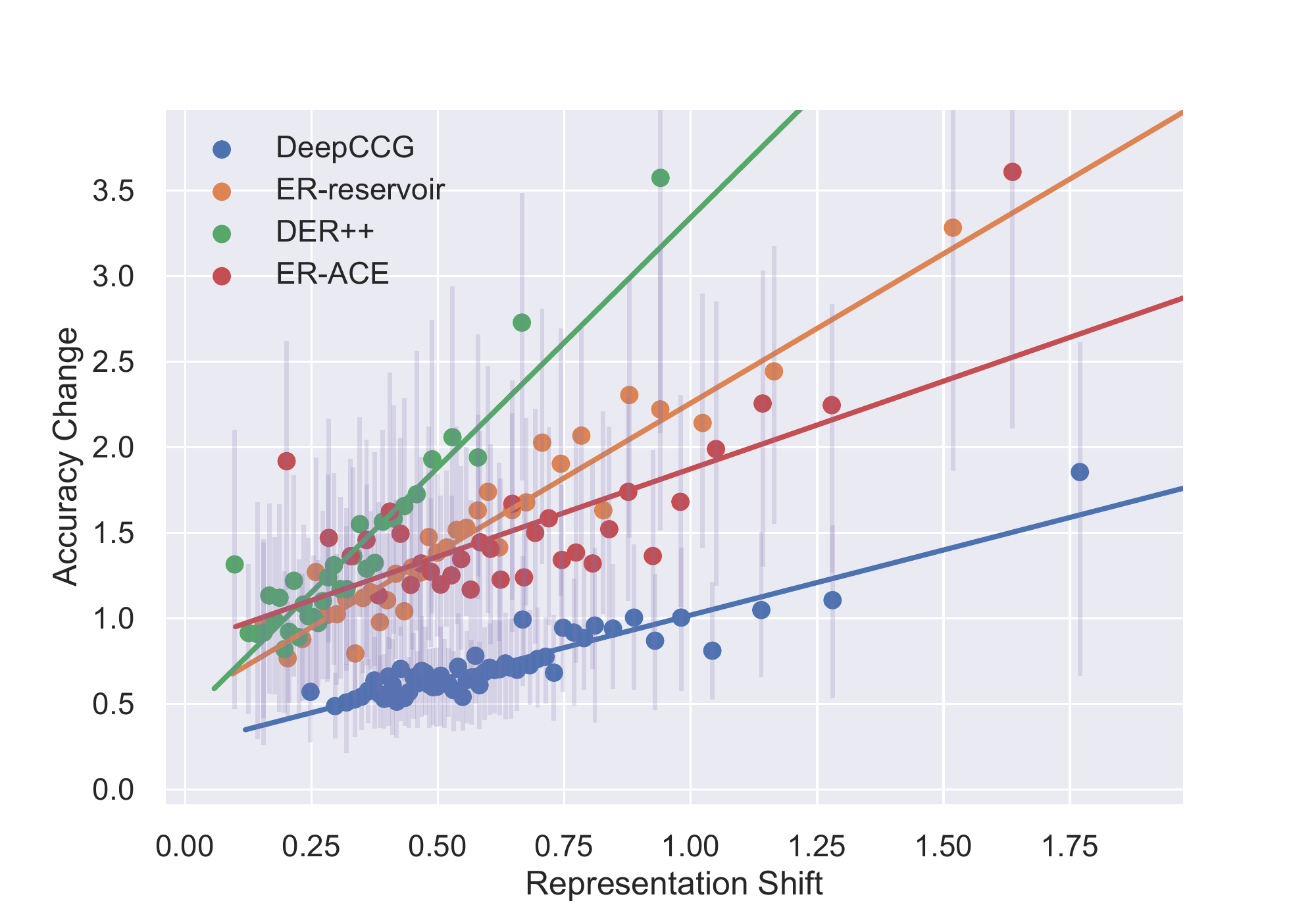}
    \caption{Binned scatter plot showing for the MiniImageNet task-incremental disjoint-tasks setting the change in accuracy against the mean change in representation after learning on a batch for the test data of the first task. The plot shows that for a given shift in representation the accuracy of DeepCCG changes the least.}
    \label{fig:repshiftplot}
    %\vspace{-2mm}
\end{figure}
\label{sec:repshift}
\begingroup
\setlength{\tabcolsep}{2.5pt} % Default value: 6pt
\begin{table*}[t]
  \caption{Results of performing an ablation on DeepCCG in the task-incremental learning scenario. We report the mean average accuracy with their standard errors across three independent runs. The results show that all components of DeepCCG are required for it to perform well.}
  \label{tab:Ablation}
  \centering
  \begin{tabular}{lllllll}
    \toprule
    & \multicolumn{2}{c}{CIFAR-10} & \multicolumn{2}{c}{CIFAR-100} & \multicolumn{2}{c}{MiniImageNet} \\
    \cmidrule(r){2-3} \cmidrule(r){4-5} \cmidrule(r){6-7}
    Method & SW & DT & SW & DT & SW & DT   \\
    \midrule
    ER-reservoir & $70.44_{\pm0.81}$ & $66.71_{\pm0.90}$ & $54.05_{\pm0.63}$ & $58.31_{\pm1.08}$ & $41.04_{\pm1.54}$ & $40.29_{\pm1.08}$ \\
    DeepCCG-reservoir & $66.63_{\pm1.47}$ & $63.97_{\pm1.06}$ & $49.95_{\pm0.743}$ & $54.76_{\pm0.08}$ & $38.76_{\pm0.65}$ & $38.14_{\pm0.10}$ \\
    DeepCCG-standardHead & $69.54_{\pm1.10}$ & $63.99_{\pm0.41}$ & $44.05_{\pm0.264}$ & $49.66_{\pm0.45}$ & $33.35_{\pm1.78}$ & $36.76_{\pm0.48}$ \\ 
    DeepCCG & $\mathbf{74.65_{\pm2.00}}$ & $\mathbf{69.29_{\pm0.89}}$ & $\mathbf{56.62_{\pm0.29}}$  & $\mathbf{60.46_{\pm0.24}}$ & $\mathbf{42.18_{\pm0.45}}$ & $\mathbf{43.04_{\pm0.64}}$ \\
    \bottomrule
  \end{tabular}
  %\vspace{-3pt}
\end{table*}
\endgroup
\textbf{Analysis of Representation Shift} \: As part of our experiments we assess how well DeepCCG's classifier update adapts to representation shift compared to other methods, showing that DeepCCG's performance is more stable with respect to representation shift. To measure representation shift, after learning on the first task and for each incoming batch of data, we recorded the mean distance over the first task's test data between its representation before and after updating on the incoming batch. Additionally, we measure the change in accuracy on first task's test data before and after updating the method on the incoming batch. We present these results for task-incremental learning on MiniImageNet in Figure~\ref{fig:repshiftplot}, displaying some of the best performing methods from our experiments. We also recorded results for other well performing methods but as they do not affect the conclusion of the experiment we do not include them in the figure to aid clarity; instead they are shown in Appendix~\ref{entSS_repshift}. Figure~\ref{fig:repshiftplot} shows that for the same amount of representation shift DeepCCG's accuracy changes less than the other methods, as its line of best fit is nearest to the horizontal axis and is the least steep. This demonstrates that the classifier of DeepCCG is more robust and better adapts to representation shift, validating DeepCCGs use of a Bayesian classifier which in one step adapts to representation shift.

\label{Ablation}
\textbf{Ablation Study} \: One way to view DeepCCG is as a relative of ER-reservoir, where the methods differ in the sample section mechanism used and the type and learning of the probabilistic classifier and embedding function. Therefore, to analyse DeepCCG, we perform an ablation, creating two adaptations of our method, DeepCCG-reservoir and DeepCCG-standardHead, by changing components of DeepCCG to what they are in ER-reservoir. DeepCCG-reservoir is when we select examples to store in memory using reservoir sampling. DeepCCG-standardHead is when we replace the class-conditional Gaussian model with a standard output head (a fully connected layer, then a softmax) and learn the embedding function using the standard ER loss, while still using DeepCCG's sample selection method. The results for the ablation are presented in Table~\ref{tab:Ablation} and show that both DeepCCG-reservoir and DeepCCG-standardHead perform much worse than DeepCCG. Therefore, we have shown that all of DeepCCG's novel components are needed in conjunction for DeepCCG to perform well. We also perform an ablation on the size of memory in Appendix~\ref{appen:mem_size}, where we show that DeepCCG performs the best over all the memory sizes tested on.

\section{CONCLUSIONS}
In this work we have demonstrated that using an empirical Bayesian procedure, DeepCCG, for online continual learning (CL) is a promising approach. The key idea of DeepCCG is to adapt the classifier in one step to representation shift, by using a Bayesian class-conditional Gaussian classifier. The correct use of such a classifier means that there is never a misalignment between the classifier and representation. Therefore, the training signal is less noisy than when there is misalignment, which happens when using previous methods. This was demonstrated in our experiments on online CL, where we show that DeepCCG is more robust to representation shift and outperforms a range of online CL methods.

\subsubsection*{Acknowledgements}
This work was kindly supported by ARM and EPSRC through an iCASE PhD scholarship. 

\bibliography{References}

%%%%%%%%%%%%%%%%%%%%%%%%%%%%%%%%%%%%%%%%%%%%%%%%%%%%%%%%%%%%
\section*{Checklist}
\begin{enumerate}

 \item For all models and algorithms presented, check if you include:
 \begin{enumerate}
   \item A clear description of the mathematical setting, assumptions, algorithm, and/or model. Yes
   \item An analysis of the properties and complexity (time, space, sample size) of any algorithm. Yes
   \item (Optional) Anonymized source code, with specification of all dependencies, including external libraries. Yes
 \end{enumerate}

 \item For any theoretical claim, check if you include:
 \begin{enumerate}
   \item Statements of the full set of assumptions of all theoretical results. Yes
   \item Complete proofs of all theoretical results. Yes
   \item Clear explanations of any assumptions. Yes     
 \end{enumerate}

 \item For all figures and tables that present empirical results, check if you include:
 \begin{enumerate}
   \item The code, data, and instructions needed to reproduce the main experimental results (either in the supplemental material or as a URL). Yes
   \item All the training details (e.g., data splits, hyperparameters, how they were chosen). Yes
   \item A clear definition of the specific measure or statistics and error bars (e.g., with respect to the random seed after running experiments multiple times). Yes
   \item A description of the computing infrastructure used. (e.g., type of GPUs, internal cluster, or cloud provider). Yes, all experiments were run using a single NVIDIA GeForce GTX 1050 GPU
 \end{enumerate}

 \item If you are using existing assets (e.g., code, data, models) or curating/releasing new assets, check if you include:
 \begin{enumerate}
   \item Citations of the creator If your work uses existing assets. Yes
   \item The license information of the assets, if applicable. Yes
   \item New assets either in the supplemental material or as a URL, if applicable. Yes
   \item Information about consent from data providers/curators. Not Applicable
   \item Discussion of sensible content if applicable, e.g., personally identifiable information or offensive content. Not Applicable
 \end{enumerate}

 \item If you used crowdsourcing or conducted research with human subjects, check if you include:
 \begin{enumerate}
   \item The full text of instructions given to participants and screenshots. Not Applicable
   \item Descriptions of potential participant risks, with links to Institutional Review Board (IRB) approvals if applicable. Not Applicable
   \item The estimated hourly wage paid to participants and the total amount spent on participant compensation. Not Applicable
 \end{enumerate}

\end{enumerate}

% Supplementary material: To improve readability, you must use a single-column format for the supplementary material.
\onecolumn
\appendix
\aistatstitle{Appendices}
%perhaps add here a section on how the classifier for class-inc learning is derrvied and why for the shifting window setting we do not use just p(y|x), marginalising over the task index. Which was beacuse the p(x,y) changes at test time as we only see a few example form the first few and last few classes in traning but at test time we see an equal amount of data from each class, so we do not use the trainig datas emprical estimate of p(y) but instead use the unifiorm dist as we know ahead of time what the test dist is (i.e. we rewieght our estimates to fit test dist). 
\section{DETAILS OF LEARNING THE EMBEDDING FUNCTION} 
\label{appen:learning}
For DeepCCG, at time $j$, to compute the update to the embedding function $f_{\phiB_j}$, when given a batch $B_j$ and a corresponding memory $M_j$, we proceed using the following steps. Firstly, we randomly sample $R_j\subset M_j$ of size $r$ from memory. Then using only the other examples in memory, $M_j/R_j$, we compute the posterior density for each class mean--- $\muB_c$, with $c \in C$:
\begin{align}
    p(\muB_c| M_j/R_j) &= p(\muB_c | M_{j, c}^Z/R_j^Z) \\
                    &= \mathcal{N}\left(\muB_y; \overline{M_{j, c}^Z/R_j^Z}, \frac{1}{m-r}\mathbf{I}\right), \label{eq:posterior}
\end{align}
where $M_{j,c}^Z=\{f_{\phiB_j}(\xB)|(\xB, y)\in M_j \land y = c \}$ are the embeddings of points in the memory buffer with class $y$. $R_j^Z$ is defined likewise and we use the notation $\overline{S}$ to denote the mean of the elements of a set $S$. Then we compute the posterior distribution of the embedded inputs $z \in B_j^Z \cup R_j^Z$ for each class $c \in C$ utilizing
\begin{align}
    p(\zB|c, M_j/R_j) &= p(\zB|c, M_{j,c}^Z/R_j^Z) \\
                  &= \int p(\zB|c,\muB_c)p(\muB_c|M_{j,c}^Z/R_j^Z) d\muB_c \\
                  &= \mathcal{N}\left(\zB; \overline{M_{j, c}^Z/R_j^Z}, (1+\frac{1}{m-r})\mathbf{I}\right). \label{eq:zMarginal}
\end{align}
Next, we compute the posterior predictive for each example $(\xB, y) \in B_j \cup R_j$ with a task identifier $t_{(\xB, y)}$, which is known for examples in the current batch and is stored by our method for examples stored in memory, and where $\zB=f_{\phiB_j}(\xB)$ using
\begin{align}
    p(y|\zB, t_{(\xB, y)}, M_j/R_j) &= \frac{p(\zB|y, M_{j,y}^Z/R_j^Z)p(y|t_{(\xB, y)})}{\sum_{c \in C}p(\zB | Y = c, M_{j,c}^Z/R_j^Z)p(Y = c | t_{(\xB, y)})} \\
                                &= \frac{p(\zB|y, M_{j,y}^Z/R_j^Z)}{\sum_{c \in C_{t_{(\xB, y)}}}p(\zB | Y = c, M_{j,c}^Z/R_j^Z)} \label{eq:yMarginal}
\end{align}

Finally, we update the embedding function by performing a gradient step using the formula  
\begin{equation}
    \phiB_{j+1} = \phiB_{j} + \eta \nabla_{\phiB} \sum_{(\xB,y) \in B_j \cup R_j} \log(p(y| \zB=f_{\phiB_j}(\xB), t_{(\xB,y)}, M_j/R_j)),
\end{equation}
which can be seen as a per-example log conditional marginal likelihood \citep{lotfi2022bayesian}.

By using Eq.~\ref{eq:zMarginal} and \ref{eq:yMarginal}, the closed form of the posterior predictive distribution for a example $(\xB, y)$ with task identifier $t_{(\xB,y)}$ and where $\zB=f_{\phiB_j}(\xB)$ is
\begin{align}
    p(y|\zB, t_{(\xB, y)}, M_j/R_j) &= \frac{p(\zB|y, M_{j,y}^Z/R_j^Z)}{\sum_{c \in C_{t_{(\xB, y)}}}p(\zB | Y = c, M_{j,c}^Z/R_j^Z)} \\
                              &= \frac{\mathcal{N}\left(\zB; \overline{M_{j, y}^Z/R_j^Z}, (1+\frac{1}{m-r})\mathbf{I} \right)}{\sum_{c \in C_{t_{(\xB,y)}}} \mathcal{N}\left(\zB; \overline{M_{j, c}^Z/R_j^Z}, (1+\frac{1}{m-r})\mathbf{I} \right)}
\end{align}

\section{PROOF OF ROBUSTNESS OF SAMPLE SELECTION MECHANISM TO i.i.d. REPRESENTATION SHIFT}
\label{ProofOfRepShift}
We show that given a shift in representation that DeepCCG's sample selection mechanism maintains the property in expectation that is minimises the KL-Divergence between the posterior over the currently accessible data and the posterior induced by the examples selected to be store in memory. The proof is as follows: Let $Z$ be the representation space and assume we are given a batch $B_{j,y}^Z $ and memory $M_{j,y}^Z$ of data points in $Z$ for a given class $y$. We define the representations shift as $\zB^*= \zB +\epsilonB_{\zB}$ where $\zB \in B_{j, y}^Z \cup M_{j,y}^Z$ and $\epsilonB_{\zB}$ is i.i.d. sampled from an arbitrary distribution with bounded mean and variance (i.e., $\mathbb{E}[\epsilonB_{\zB}] < \infty$ and $\Var(\epsilonB_{\zB}) < \infty)$). Furthermore, define $\etaB_{\zB} = \epsilonB_{\zB} - \mathbb{E}[\epsilonB_z]$, hence it is $\epsilonB$ shifted to have zero mean. Additionally, let $B_{j, y}^{Z^*}=\{\zB^*=\zB+\epsilonB_{\zB}| \zB \in B_{j, y}^Z\}$ and define $M^{Z^*}_{j, y}$ and $M_y^{\prime Z^{*}}$ likewise. Therefore we have that,
\begingroup
\allowdisplaybreaks
\begin{align}
    &\mathbb{E}[\mathrm{KL}(p(\muB_y|D^{Z^*}_{j, y},M^{Z^*}_{j, y}) \lVert p(\muB_y|M^{\prime Z^*}_y)))] = \mathbb{E}[\lVert \overline{B_{j, y}^{Z^*} \cup M_{j, y}^{Z^*}}-\overline{M_y^{\prime Z^{*}}} \rVert^2_2] \\
    &= \mathbb{E}[ \lVert \frac{1}{|B_{j, y}^{Z^*} \cup M_{j, y}^{Z^*}|} \sum_{\zB^* \in B_{j, y}^{Z^*} \cup M_{j, y}^{Z^*}} \zB^* - \frac{1}{|M_y^{\prime Z^{*}}|} \sum_{\zB^* \in M_y^{\prime Z^{*}}} \zB^* \rVert^2_2] \\
    &= \sum_{k = 0}^{d} \mathbb{E}[(\frac{1}{|B_{j, y}^{Z^*} \cup M_{j, y}^{Z^*}|} \sum_{\zB^* \in B_{j, y}^{Z^*} \cup M_{j, y}^{Z^*}} z^{*}_k - \frac{1}{|M_y^{\prime Z^{*}}|} \sum_{\zB^* \in M_y^{\prime Z^{*}}} z^{*}_k)^2] \\
    &= \sum_{k = 0}^{d} \mathbb{E}[(\frac{1}{|B_{j, y}^{Z} \cup M_{j, y}^{Z}|} \sum_{\zB \in B_{j, y}^{Z} \cup M_{j, y}^{Z}} (z_k + \mathbb{E}[\epsilon_{z,k}] +\eta_{z,k}) - \frac{1}{|M_y^{\prime Z}|} \sum_{\zB \in M_y^{\prime Z}} (z_k + \mathbb{E}[\epsilon_{z, k}] +\eta_{z, k}))^2] \\
    &= \sum_{k = 0}^{d} \mathbb{E}[(\frac{1}{|B_{j, y}^{Z} \cup M_{j, y}^{Z}|} \sum_{\zB \in B_{j, y}^{Z} \cup M_{j, y}^{Z}} (z_k +\eta_{z, k}) - \frac{1}{|M_y^{\prime Z}|} \sum_{\zB \in M_y^{\prime Z}} (z_k + \eta_{z, k}) \\ 
    & \; + (\frac{1}{|B_{j, y}^{Z} \cup M_{j, y}^{Z}|} \sum_{\zB \in B_{j, y}^{Z} \cup M_{j, y}^{Z}} \mathbb{E}[\epsilonB_{z, k}] - \frac{1}{|M_y^{\prime Z}|} \sum_{\zB \in M_y^{\prime Z}} \mathbb{E}[\epsilonB_{z, k}]))^2] \\
    &= \sum_{k = 0}^{d} \mathbb{E}[(\frac{1}{|B_{j, y}^{Z} \cup M_{j, y}^{Z}|} \sum_{\zB \in B_{j, y}^{Z} \cup M_{j, y}^{Z}} (z_k +\eta_{z, k}) - \frac{1}{|M_y^{\prime Z}|} \sum_{\zB \in M_y^{\prime Z}} (z_k + \eta_{z, k}))^2] \text{\; (as } \epsilon_{z,k} \text{ are i.i.d.)} \\
    &= \sum_{k = 0}^{d} \mathbb{E}[(\frac{1}{|B_{j, y}^{Z} \cup M_{j, y}^{Z}|} \sum_{\zB \in B_{j, y}^{Z} \cup M_{j, y}^{Z}} z_k - \frac{1}{|M_y^{\prime Z}|} \sum_{\zB \in M_y^{\prime Z}} z_k \\
    & + \frac{1}{|B_{j, y}^{Z} \cup M_{j, y}^{Z}|} \sum_{\zB \in B_{j, y}^{Z} \cup M_{j, y}^{Z}} \eta_{z, k} - \frac{1}{|M_y^{\prime Z}|} \sum_{\zB \in M_y^{\prime Z}} \eta_{z, k})^2] \\
    &= \sum_{k = 0}^{d} \mathbb{E}[(\frac{1}{|B_{j, y}^{Z} \cup M_{j, y}^{Z}|} \sum_{\zB \in B_{j, y}^{Z} \cup M_{j, y}^{Z}} z_k - \frac{1}{|M_y^{\prime Z}|} \sum_{\zB \in M_y^{\prime Z}} z_k)^2] \\
    & + 2\mathbb{E}[(\frac{1}{|B_{j, y}^{Z} \cup M_{j, y}^{Z}|} \sum_{\zB \in B_{j, y}^{Z} \cup M_{j, y}^{Z}} z_k - \frac{1}{|M_y^{\prime Z}|} \sum_{\zB \in M_y^{\prime Z}} z_k)(\frac{1}{|B_{j, y}^{Z} \cup M_{j, y}^{Z}|} \sum_{\zB \in B_{j, y}^{Z} \cup M_{j, y}^{Z}} \eta_{z, k} - \frac{1}{|M_y^{\prime Z}|} \sum_{\zB \in M_y^{\prime Z}} \eta_{z, k})] \\ 
    & + \mathbb{E}[(\frac{1}{|B_{j, y}^{Z} \cup M_{j, y}^{Z}|} \sum_{\zB \in B_{j, y}^{Z} \cup M_{j, y}^{Z}} \eta_{z, k} - \frac{1}{|M_y^{\prime Z}|} \sum_{\zB \in M_y^{\prime Z}} \eta_{z, k})^2] \\
    &= \sum_{k = 0}^{d} (\frac{1}{|B_{j, y}^{Z} \cup M_{j, y}^{Z}|} \sum_{\zB \in B_{j, y}^{Z} \cup M_{j, y}^{Z}} z_k - \frac{1}{|M_y^{\prime Z}|} \sum_{\zB \in M_y^{\prime Z}} z_k)^2 \\
    & + \sum_{k = 0}^{d} \mathbb{E}[(\frac{1}{|B_{j, y}^{Z} \cup M_{j, y}^{Z}|} \sum_{\zB \in B_{j, y}^{Z} \cup M_{j, y}^{Z}} \eta_{z, k} - \frac{1}{|M_y^{\prime Z}|} \sum_{\zB \in M_y^{\prime Z}} \eta_{z, k})^2] \text{\; (as } \mathbb{E}[\eta_{z,k}] = 0 \text{)} \\
    &= \lVert \overline{B_{j, y}^{Z} \cup M_{j, y}^{Z}}-\overline{M_y^{\prime Z}} \rVert^2_2  + \sum_{k = 0}^{d} \left( \mathbb{E}[(\frac{1}{|B_{j, y}^{Z} \cup M_{j, y}^{Z}|} \sum_{\zB \in B_{j, y}^{Z} \cup M_{j, y}^{Z}} \eta_{z, k})^2] + \mathbb{E}[(\frac{1}{|M_y^{\prime Z}|} \sum_{\zB \in M_y^{\prime Z}} \eta_{z, k})^2] \right. \\
    &  - 2\mathbb{E}[(\left.\frac{1}{|B_{j, y}^{Z} \cup M_{j, y}^{Z}|} \sum_{\zB \in B_{j, y}^{Z} \cup M_{j, y}^{Z}} \eta_{z, k})(\frac{1}{|M_y^{\prime Z}|} \sum_{\zB \in M_y^{\prime Z}} \eta_{z, k})] \right) \\
    &= \lVert \overline{B_{j, y}^{Z} \cup M_{j, y}^{Z}}-\overline{M_y^{\prime Z}} \rVert^2_2 + \sum_{k = 0}^{d} \left(\frac{\Var(\epsilon_k)}{|B_{j, y}^{Z} \cup M_{j, y}^{Z}|} + \frac{\Var(\epsilon_k)}{|M_y^{\prime Z}|} \right. \\
    &  - \left.\frac{2}{|B_{j, y}^{Z} \cup M_{j, y}^{Z}||M_y^{\prime Z}|}\sum_{\zB \in B_{j, y}^{Z} \cup M_{j, y}^{Z}}\sum_{\zB' \in M_y^{\prime Z}}\mathbb{E}[\eta_{z, k}\eta_{z', k}]\right)  \text{\; (using formula for variance of means)} \\
    &= \lVert \overline{B_{j, y}^{Z} \cup M_{j, y}^{Z}}-\overline{M_y^{\prime Z}} \rVert^2_2 +(\frac{1}{|B_{j, y}^{Z} \cup M_{j, y}^{Z}|} + \frac{1}{|M_y^{\prime Z}|}) \sum_{k = 0}^{d} \Var(\epsilon_k) \\
    & - \frac{2}{|B_{j, y}^{Z} \cup M_{j, y}^{Z}||M_y^{\prime Z}|} \sum_{k = 0}^{d}  \sum_{\zB \in M_y^{\prime Z}} \mathbb{E}[\eta_{z, k}^2] \\
    &= \lVert \overline{B_{j, y}^{Z} \cup M_{j, y}^{Z}}-\overline{M_y^{\prime Z}} \rVert^2_2 +(\frac{1}{|B_{j, y}^{Z} \cup M_{j, y}^{Z}|} + \frac{1}{|M_y^{\prime Z}|}) \sum_{k = 0}^{d} \Var(\epsilon_k) - \frac{2}{|B_{j, y}^{Z} \cup M_{j, y}^{Z}|} \sum_{k = 0}^{d} \Var(\epsilon_k) \\
    &= \lVert \overline{B_{j, y}^{Z} \cup M_{j, y}^{Z}}-\overline{M_y^{\prime Z}} \rVert^2_2 +(\frac{1}{|M_y^{\prime Z}|} - \frac{1}{|B_{j, y}^{Z} \cup M_{j, y}^{Z}|}) \sum_{k = 0}^{d} \Var(\epsilon_k) 
\end{align} 
\endgroup
which completes the proof and where in the main paper we define $\nu = \sum_{k = 0}^{d} \Var(\epsilon_k)$.

\FloatBarrier
\section{TABLE OF CONTINUAL LEARNING TERMINOLOGY}
\label{appen:CLterms}
\begin{table}[h!]
    \centering
    \caption{Definition of continual learning terms used in the text.}
    \begin{tabular}{p{0.3\textwidth}p{0.65\textwidth}}
         \toprule
         \textbf{Term} & \textbf{Definition}  \\
         \midrule
         \textbf{Online continual learning} & A continual learning setup where the learner sees the data batch by batch and cannot revisit previous batches. Each batch is generated by a task $t \in T$, and when training the learner is told what task that is. \\
         \midrule
         \textbf{Task} & A task $t \in T$ is a data generator, which generates data (i.e. data instances $\xB \in X$ and labels $y \in C$) from a given subset of the classes $C_t \subseteq C$.\\
         \midrule
         \textbf{Task identifier} & States for some example $(\xB, y)$ (or batch of examples) what task it was generated from. \\
         \midrule
         \textbf{Disjoint tasks setting} & A setting where no two tasks generate data from the same class. In other words for any two task $t$ and $t'$ we have that $C_{t} \cap C_{t'} = \emptyset$.\\
         \midrule
         \textbf{Shifting window setting} & A setting where we fix an ordering of the classes $c_1,\ldots, c_k$ and define the $i$th task to generate data from the classes $c_i,\ldots,c_{i+l}$, where $l$ is a free parameter of setting called the window length. \\
         \midrule
         \textbf{Task-incremental learning} & A scenario where at test time the learner has access to the task identifiers of the data instances in the test set. \\
         \midrule
         \textbf{Class-incremental learning} & A scenario where at test time the learner does not have access to task identifiers and so classifies across all tasks/classes seen.\\         
         \bottomrule
    \end{tabular}
\end{table}
\FloatBarrier

\section{DESCRIPTION OF METHODS USED FOR COMPARISON}
\label{appen:methods}
In our experiements we compared DeepCCG to the following continual learning methods: EWC \citep{huszar2018note, kirkpatrick2017overcoming}, PackNet \citep{mallya2018packnet}, ER-Reservoir \citep{Chaudhry2020Continual}, A-GEM \citep{Chaudhry2019Efficient}, EntropySS \citep{Wiewel2021Entropy}, GSS \citep{Aljundi2019Gradient}, ER-ACE \citep{caccia2021new}, DER++ \citep{buzzega2020dark}, ESMER \citep{sarfraz2022error} and iCaRL \citep{rebuffi2017icarl}. Below we give a short description of each method.

\begin{itemize}
    \item[] \textbf{EWC} is a parameter regularisation approach which uses an $l_2$ regulariser to ensure the parameters do not move to far away from the weighted average of the parameters fitted for the previous tasks. Additionally, this $l_2$ regulariser is weighted using the Fisher information of the fitted parameters on previous tasks.    

    \item[] \textbf{PackNet} is a parameter isolation method which freezes a certain proportion of the filters/nodes after learning on a task. This means that for future tasks learning takes place only on the remaining part of the neural network which is not frozen, preventing the forgetting of the subnetworks used for the previous tasks. To select which filters are frozen for a task PackNet uses a filter pruning method.   

    \item[] \textbf{ER-Reservoir} is the standard replay method. It stores previously seen examples in a replay buffer and in training ``replays" a random batch of the stored examples by training on them along side the examples of the current batch in the data stream. Also, it selects what examples to store in memory using reservoir sampling \citep{vitter1985random}.

    \item[] \textbf{A-GEM} is a replay method which prevents the average loss on examples stored in the memory buffer from increasing. This is achieved by projecting the proposed weight update vector (i.e. the gradient of the loss w.r.t. the weights), such that the above constraint is satisfied while minimally changing the update vector.      

    \item[] \textbf{EntropySS} replaces the sample selection method of ER-Reservoir with one which chooses which examples to store in memory by trying to ensure the entropy of the stored examples is maximised. 

    \item[] \textbf{GSS} replaces the sample selection method of ER-Reservoir with one which chooses which examples to store in memory by maximising the variance of gradient values of the loss for the stored examples.

    \item[] \textbf{ER-ACE} is a replay method which uses pseudo task-identifiers such that previous tasks are seen to be merged into a single task.

    \item[] \textbf{DER++} is a replay method which in addition to using standard replay adds another regularisation term which prevents the logits of previous examples straying from the values they took when they were given to the network for the first time. 

    \item[] \textbf{ESMER} is a replay method which uses knowledge distillation to an exponentially moving average of the network to regularise updates.

    \item[] \textbf{iCaRL} is a replay method which uses a distillation regulariser and per-class sigmoids when learning. Additionally, at test time it uses a nearest means classifier, using the examples stored in memory to form the per-class means. Importantly, we use the adaptation of iCaRL given in \citet{buzzega2020dark}, such that it can be used in task incremental learning. 
    
\end{itemize}

\section{ADDITIONAL EXPERIMENTAL DETAILS}
\label{appen:addExpDetails}
\footnote{Code for DeepCCG is available at \href{https://github.com/Tlee43/DeepCCG}{https://github.com/Tlee43/DeepCCG}.}In addition to the experimental details mentioned in the main body of the paper, there are a few more details to mention. First, for methods with hyperparameters we performed a grid search using the same experimental set up as the real experiments and $10\%$ of the training data as validation data. This led to the selection of the hyperparameters of EWC and PackNet shown in Tables~\ref{hyperparams1} and \ref{hyperparams2}. For DER++ and ESMER we tune the hyperparams on CIFAR-100 and then use the same hyperparams across all datasets. The hyperparameters selected for DER++ were $\alpha=0.5$ and $\beta=1.0$; and for ESMER were $\alpha_{EMA}=0.999$, $\alpha_{l}=0.99$, $\beta=2.5$, $\gamma = 0.2$ and $r=0.7$. Second, we use the same learning rate of $0.1$ and the standard gradient decent optimiser for all methods. These were chosen by looking at the commonly selected values in previous work \citep{Mirzadeh2020Understanding} and were shown to be performative for all methods tested. Third, for the multi-task upper bound we train using two epochs in the task-incremental scenario and three epochs in the class-incremental scenario, to more fairly upper bound the performance of the methods. Last, as ESMER stores both an exponential moving average of the model and examples in memory, we ensure it uses roughly the same amount of memory as other methods by reducing the number of examples it stores. This means that for experiments on CIFAR-100 and MiniImageNet we reduce ESMER's memory size by 400 examples (the rough number of examples that take up the same space as a copy of the model) and we do not reduce the memory size for CIFAR-10 as it is already less than 400 examples.
\begin{table}[h]
  \caption{Values selected for the hyperparameters of EWC and PackNet in the task-incremental learning experiments, which are the regularisation coefficient and the percentage of available filters to be used per task, respectively. SW stands for shifting window and DT stands for disjoint tasks.}
  \label{hyperparams1}
  \centering
  \begin{tabular}{lllllll}
    \toprule
    & \multicolumn{2}{c}{CIFAR-10} & \multicolumn{2}{c}{CIFAR-100} & \multicolumn{2}{c}{MiniImageNet} \\
    \cmidrule(r){2-3} \cmidrule(r){4-5} \cmidrule(r){6-7}
    Method & SW & DT & SW & DT & SW & DT  \\
    \midrule
    EWC & $1$ & $6$ & $6$ & $6$ &  $2$ & $9$   \\
    PackNet & $0.3$ & $0.3$ & $0.1$ & $0.1$ & $0.05$ & $0.2$  \\
    \bottomrule
  \end{tabular}
\end{table}
\begin{table}[h]
  \caption{Values selected for the hyperparameters of EWC in the class-incremental learning experiments, which is the regularisation coefficient. SW stands for shifting window and DT stands for disjoint tasks.}
  \label{hyperparams2}
  \centering
  \begin{tabular}{lllllll}
    \toprule
    & \multicolumn{2}{c}{CIFAR-10} & \multicolumn{2}{c}{CIFAR-100} & \multicolumn{2}{c}{MiniImageNet} \\
    \cmidrule(r){2-3} \cmidrule(r){4-5} \cmidrule(r){6-7}
    Method & SW & DT & SW & DT & SW & DT  \\
    \midrule
    EWC & $1$ & $1$ & $6$ & $1$ &  $6$ & $1$   \\
    \bottomrule
  \end{tabular}
\end{table}

\section{ADDITIONAL RESULTS}
\subsection{Results on Using Different Memory Sizes}
\label{appen:mem_size}
\begin{table*}[h]
  \caption{Results of experiments looking at the effect of memory buffer size $m$ for the replay methods tested, using the shifting window setting on CIFAR-100 in task-incremental learning scenario. We report mean average accuracy with their standard errors across three independent runs.}
  \label{tab:Mem_size}
  \centering
  \begin{tabular}{lllll}
    \toprule
    Method & m=750 & m=1000 & m=1250 \\
    \midrule
    ER-reservoir &  $53.17_{\pm0.656}$ & $54.05_{\pm0.626}$ & $55.22_{\pm0.592}$ \\
    A-GEM & $27.18_{\pm1.091}$ & $29.01_{\pm1.449}$ & $27.87_{\pm0.172}$ \\
    EntropySS & $50.52_{\pm0.725}$ & $51.80_{\pm0.700}$ & $53.34_{\pm0.372}$ \\
    GSS & $47.15_{\pm0.766}$ & $48.20_{\pm0.332}$ & $46.18_{\pm0.341}$ \\
    DeepCCG (ours) &  $\mathbf{53.42_{\pm0.460}}$ & $\mathbf{56.62_{\pm0.288}}$ & $\mathbf{57.98_{\pm0.514}}$ \\
    \bottomrule
  \end{tabular}
\end{table*}
One additional useful experiment is looking at the relationship between performance and the size of memory used for DeepCCG. Therefore, in Table \ref{tab:Mem_size} we show the performance of DeepCCG and other replay methods compared against with varying memory size. Table \ref{tab:Mem_size} shows that when the memory size is increased to $m=1250$, DeepCCG has an improvement in average accuracy relative to other methods, as it achieves $2.76\%$ better than any other method for $m=1250$, while for $m=1000$ it achieves $2.57\%$ better than any other method, a $0.19\%$ improvement. When the memory size is decreased to $m=750$, we see that DeepCCG's performance drops more than other methods as it is only $0.25\%$ better than other methods in this case. Therefore, our experiments show that compared to other replay methods DeepCCG's performance increases the most when $m$ increases and that for small buffer sizes DeepCCG performs less well, potentially due to the fact that the examples in the memory buffer are used to infer the posterior over the means of the per-class Gaussians and so the method needs a given amount of examples to specify the means well. We also note that in our experiments the performance ranking of the methods does not change with $m$.

%\vspace{5cm}
\subsection{Additional Representation Shift Results}
\label{entSS_repshift}
\begin{figure*}[h]
    \centering
    \includegraphics[trim={1.1cm 0.0cm 2cm 1.7cm}, clip, scale=0.4]{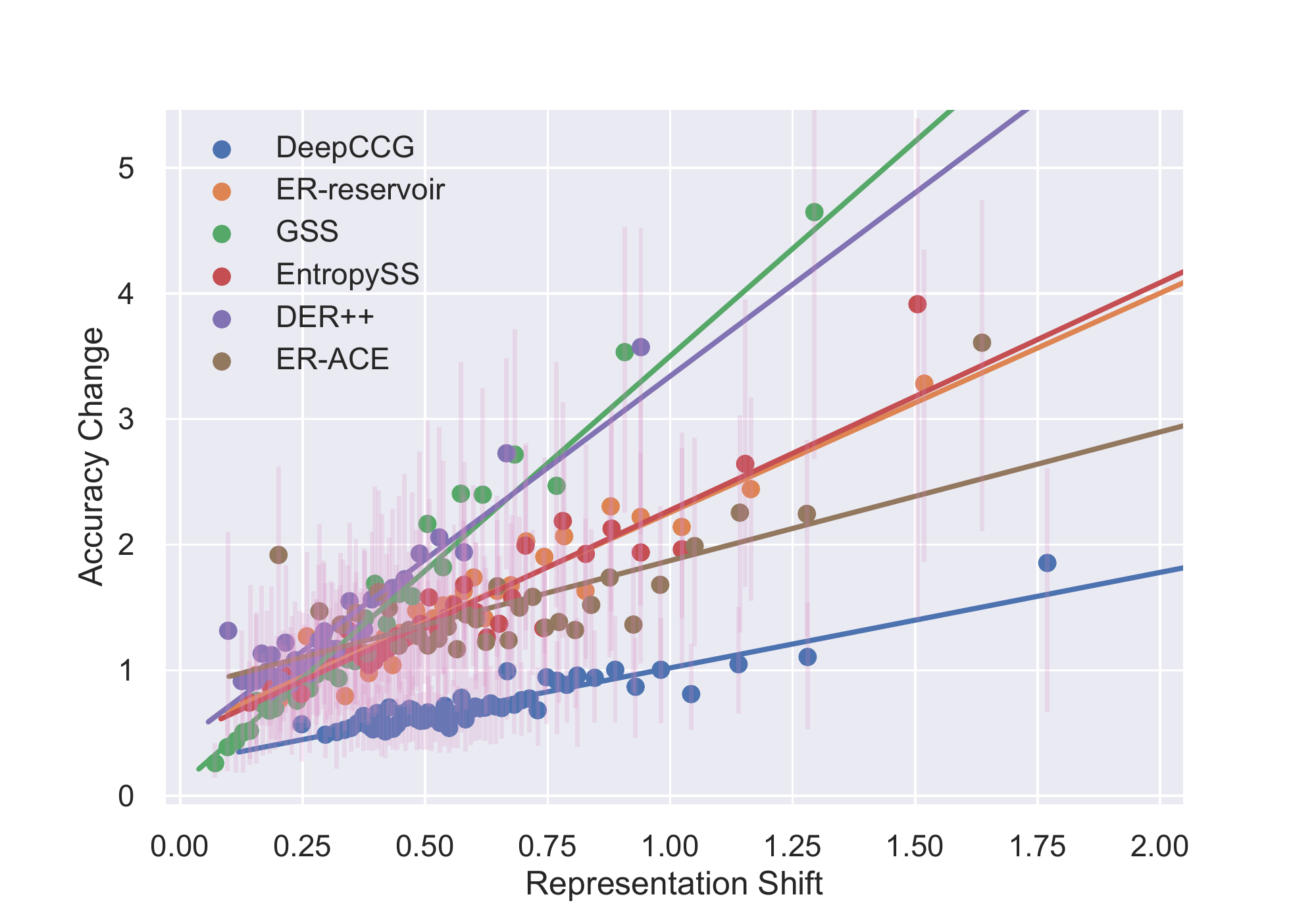}
    \caption{Binned scatter plot showing for the MiniImageNet disjoint tasks setting the change in accuracy against the mean change in representation after learning on a batch for the test data of the first task.}
\end{figure*}

\end{document}